\documentclass[runningheads]{llncs}
\usepackage[T1]{fontenc}
\author{Tom Pieper\inst{1}\orcidID{0009-0001-8978-6577} \and
Mohamad Ballout\inst{1} \and
Ulf Krumnack\inst{1}\orcidID{0000-0003-1976-8186} \and
Gunther Heidemann\inst{1} \and 
Kai-Uwe Kühnberger\inst{1}\orcidID{0000-0003-1626-0598}}

\institute{Institute of Cognitive Science, Osnabrueck University, 49069 Osnabrück, Germany \\
\email{tomlpieper@gmail.com}}

\usepackage{graphicx}
\usepackage{multirow}
\usepackage{wrapfig}
\usepackage{subcaption}
\usepackage{amsmath}

\usepackage{xcolor}
\usepackage{listings}
\definecolor{asparagus}{HTML}{678d58}
\definecolor{metal}{HTML}{242f40}

\lstdefinestyle{mystyle}{
    language=Python,
    showspaces=false,
    showtabs=false,
    breaklines=true,
    showstringspaces=false,
    stringstyle=\color{metal},
    basicstyle=\ttfamily\footnotesize,
    commentstyle=\color{gray},
    numbers=left, 
    numberstyle=\tiny\color{gray}, 
    stepnumber=1, 
    numbersep=5pt, 
    xleftmargin=1em
}
\begin{document}
\title{Enhancing Small Language Models via ChatGPT and Dataset Augmentation}
\titlerunning{Enhancing SLM via ChatGPT and Dataset Augmentation}
%

%
\authorrunning{T. Pieper et al.}
%
%
\maketitle              
\begin{abstract}
This paper explores the enhancement of small language models through strategic dataset augmentation via ChatGPT-3.5-Turbo, in the domain of Natural Language Inference (NLI). By employing knowledge distillation-based techniques and synthetic dataset augmentation, we aim to bridge the performance gap between large language models (LLMs) and small language models (SLMs) without the immense cost of human annotation. 
Our methods involve two forms of rationale generation--information extraction and informed reasoning--to enrich the ANLI dataset. We then fine-tune T5-Small on these augmented datasets, evaluating its performance against an established benchmark.
Our findings reveal that the incorporation of synthetic rationales significantly improves the model's ability to comprehend natural language, leading to 1.3\% and 2.3\% higher classification accuracy, respectively, on the ANLI dataset, demonstrating the potential of leveraging LLMs for dataset augmentation. This approach not only enhances the performance of smaller models on complex tasks but also introduces a cost-effective method for fine-tuning smaller language models.
By advancing our understanding of knowledge distillation and fine-tuning strategies, this work contributes to the ongoing effort to create more capable and efficient NLP systems.

\end{abstract}
\section{Introduction}
In the rapidly evolving field of artificial intelligence, Natural Language Processing (NLP) stands out for its capability to bridge the gap between human-machine interaction and communication via natural language. It makes everyday tasks possible without special encoding or structuring of every bit of data. These tasks include powering virtual assistants, machine translators, and extracting information. The recent emergence of Large Language Models (LLMs) like GPT-3 \cite{Brown2020GPT3} has revolutionized the generative and comprehension abilities of machines regarding natural language, enabling human-like communication.
This increased user-friendly accessibility sparks a renewed interest in artificial intelligence, accelerating advancements in this field. 
While the capabilities of LLMs consistently grew in a variety of tasks, so did the models' sizes, making use of an increasing number of parameters. An apparent challenge in the field of NLP, therefore, arises in optimizing smaller models for tasks that their large counterparts may be able to solve, but the amount of computational power needed for a specific task does not justify the cost of the model.
This paper explores a combination of these models using their inherent strengths to complement their weaknesses through knowledge distillation. In contrast to classical approaches like \cite{ballout2024its}, we aim to improve a small language model's performance beyond its standard capabilities on a specific task by fine-tuning it on rationales generated by a large, more capable model, i.e., via dataset augmentation. 

As a result, our core focus lies in augmenting a Natural Language Inference (NLI) dataset by strategically prompting ChatGPT-3.5-Turbo with parts of the ANLI dataset using two different techniques. The first rationale generated is based on the principle of information extraction. It comprises the essence of the premise by answering the 5 W-Questions (Who, What, When, Where, Why). 
The second version is based on informed reasoning and consists of a free-text rationale generated by ChatGPT-3.5-Turbo that explains and justifies the label classification of each premise-hypothesis pair.
Both datasets are tested using a variety of different T5-Small configurations in an Input $\to$ (Output $+$ Rationale) fashion utilizing a custom split loss for training. 
The baseline is set by a T5-Small in an Input $\to$ Output fashion (standard practice for classification).

Based on the foundational concept introduced in Chapter~\ref{RW}, this paper proves the benefits of augmenting an NLI dataset without human annotation and concludes the inherent natures and effects of rationale types.

\section{Related Work} \label{RW}
\subsection{NLI}
The interest in research concerning NLP has accelerated in recent years due to increased pathways of capability and applicability of LLMs built on attention-mechanism \cite{vaswani2023attention} based transformer models that elevated the comprehension and generation of natural language manifold.\\
The capacity for linguistic inference, defined as discerning an interlocutor's implied meaning beyond literal expressions and generating contextually relevant discourse, is essential for facilitating communication \cite{bambini2023pragmaticprofileGPT}. The problem of NLI can hence be seen as one of the main challenges when \textit{bridging the gap} between human and machine communication.

More recent research focused on different influences, active human-model interactions, and improving the quality of the datasets to improve overall downstream task performance. One promising approach is to include rationales and different forms of reasoning in the datasets and, hence, the fine-tuning or training process. These annotations are often formulated and generated by humans \cite{camburu2018esnli,nie2020adversarial}, and especially benefit smaller models like T5 \cite{raffel2023exploring}.
LLMs with hundreds of billions of parameters perform quite well on these NLI datasets, exhibiting a decent semantic understanding due to their size and enormous training data. Smaller language models with only millions of parameters struggle more with these abstract concepts.

\subsection{Knowledge Distillation}
Attempting to enable these smaller, more applicable models to comprehend natural language more profoundly opens up a new area of possible research. The concept of knowledge distillation (originally called model compression \cite{modelcompression2006}) is used by a variety of authors to improve the performance of smaller student models by transferring (distilling) knowledge from an ensemble of models \cite{hinton2015distilling} or in more recent cases from a magnitudes stronger teacher model \cite{hsieh2023distilling}. In the latter case a transfer function \cite{wang2022kd} is utilized. In most cases, parameters or complete layers of the teacher are embedded into the student model \cite{ballout2024its}. 

A similar and rather thorough approach to transfer knowledge to smaller models through reasoning is given by \cite{ho2023large} that proposes a further developed Chain-of-thought method \cite{wei2023chainofthought} for knowledge distillation. 
In the same fashion, \cite{li2022explanations} elicit increased reasoning abilities in small models utilizing the capabilities of GPT-3 while also generating free-text high-quality explanations.

Another approach to NLP problem-solving using only text-to-text models is introduced in \cite{raffel2023exploring}, where capacities of transfer learning from data-rich environments to different downstream tasks are used.

Different concepts of rationales can be applied to bolster a model's performance while fine-tuning an NLI dataset.
While extractive explanations are fairly limited in their expressiveness and chance to lay open the model decisions, they are more straightforward to measure with tools/benchmarks like those proposed by \cite{DeYoung2019ERASER}. At the same time, Free-Text Rationales can provide more insight into the background of the model's decision. Still, especially in model-generated Free-Text rationale, it is hard to check the relevance and coherence of these rationales without human supervision \cite{narang2020wt5}. A detailed view into the reasons and ways for models benefiting from specific explanations is further tested and given by~\cite{hase-bansal-2022-models}.

\subsection{Synthetic Dataset Augmentation}
Augmenting datasets is a strategy frequently employed in computer vision tasks aimed at boosting models' generalization capabilities. However, the augmentation process benefits all types of models by providing them with a training dataset that is diverse, robust, and comprehensive. This process is often done by independent workers from Amazon Mechanical Turk as done by the authors of the \textit{CoS-E} \cite{rajani2019explain}, the \textit{e-SNLI} \cite{camburu2018esnli} datasets that used these human explanations to modify the original datasets \textit{SNLI} \cite{maccartney-manning-2008-modeling} and\textit{ CQA }proposed by \cite{talmor2019commonsenseqa}.  Similar to this, the authors of \textit{FLUTE} \cite{Chakrabarty2022FLUTE} and  \textit{ANLI} \cite{nie2020adversarial} made use of these portals to hire workers.

Considering utilizing the recent, strong LLMs for the augmentation of NLU datasets hence provides a challenging yet rewarding attempt to overcome these immense costs by utilising the LLMS increased Few- and Zero-Shot-Performance on challenging tasks.

Introducing the ``worker and ai collaboration'' \cite{liu-etal-2022-wanli} utilizes the diversity of the output GPT-3 can provide to create a new non-adversarial approach to human-machine interaction for the dataset.

The recent publications of the frameworks \textit{ZeroGen} \cite{Ye2022zeroGen}, \textit{SuperGen} proposed in the same year by  \cite{meng2022generating} and the iterative \textit{Synthesise-step-by-step framework (S3)} by \cite{wang2023lets} completely remove the need for human supervision for the generation of training data in the field of NLP, more precisely natural language understanding (NLU) and NLI.
The same approach of indirect transfer of the teacher model's knowledge through dataset synthesis will be taken in this paper.

\section{Methods}
\begin{figure}[t]
     \centering
     \begin{subfigure}[b]{0.49\textwidth}
         \centering
         \includegraphics[width=\textwidth]{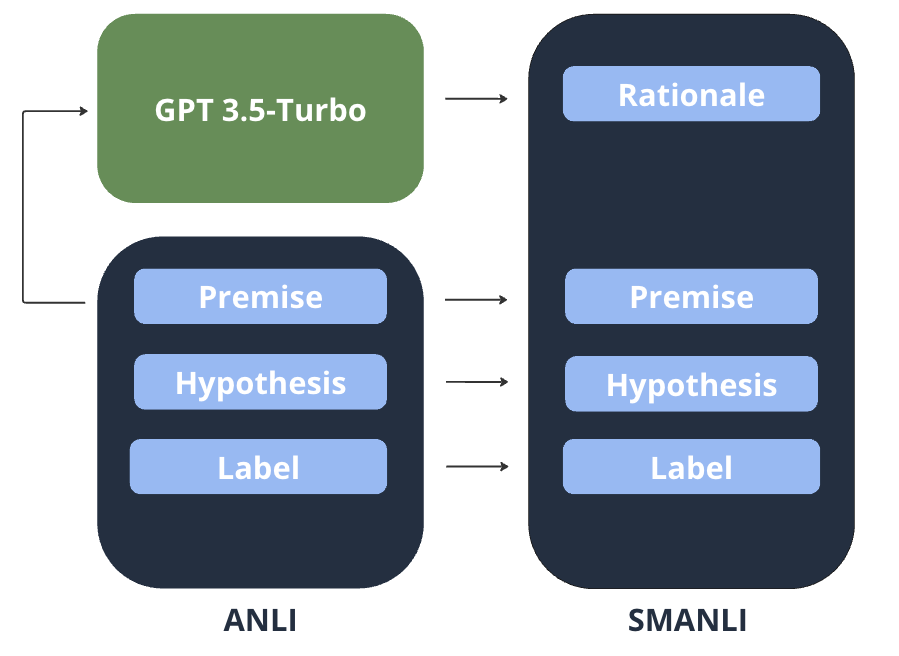}
         \caption{Generation process of \textit{SMANLI}.}
         \label{fig:GenSMANLI}
     \end{subfigure}
     \hfill
     \begin{subfigure}[b]{0.49\textwidth}
         \centering
         \includegraphics[width=\textwidth]{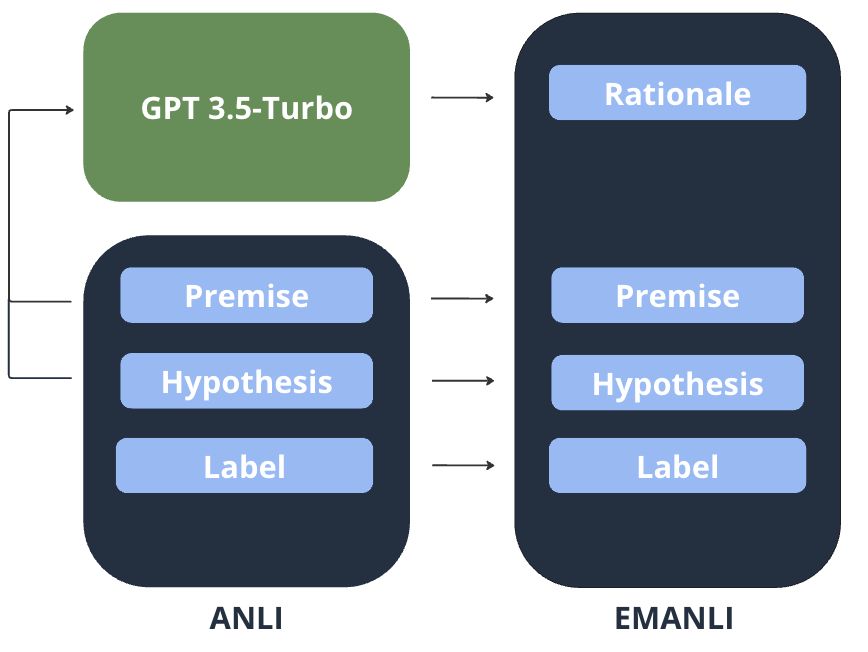}
         \caption{Generation process of \textit{EMANLI}.}
         \label{fig:GenEMANLI}
     \end{subfigure}

        \caption{Generation of the modified datasets.}
        \label{fig:datasetGen}
\end{figure}

In the following, we will provide the experimental design chosen to prove and evaluate the positive effect of knowledge distillation via dataset augmentation of the ANLI dataset \cite{nie2020adversarial} on fine-tuning a T5-Small model \cite{raffel2023exploring} using the Hugginface API \cite{wolf2020huggingfaces}. 
The utilization of ChatGPT-3.5-Turbo to augment ANLI \cite{nie2020adversarial} can be found in Figure~\ref{fig:datasetGen}.
The full code and datasets are provided in our public Github repository \cite{githubRepo}.

\subsubsection{Summarised Modified ANLI (SMANLI)}
The for SMANLI generated rationale is the distilled essential information extracted from the hypothesis of each data point based on the  ``5 Ws'' approach (Who, What, When, Where, Why). Based on the information extraction principle, this strategy provides a simplified yet comprehensive view of the relevant text, training the SLM to extract essential information for elevated comprehension. For each datapoint of ANLI to be augmented ChatPGT-3.5-Turbo was prompted as follows:

\begin{lstlisting}[language=Python]
"Answer the 5 W questions about the following text with max 10 words per question: {premise}
Who:, What: When, Where:, Why:"
\end{lstlisting}
The already low cost of this augmentation process could be further reduced by grouping the hypotheses and reducing the 16,946 prompts to 12\% of their original amount.

\subsubsection{Explained Modified ANLI (EMANLI)}
The EMNALI dataset results from a more nuanced, encompassingly informed way of prompting the teacher model. The generated rationales consist of free-text rationales that justify the label classification for each premise-hypothesis pair. This strategy enhances the dataset with increased semantic depth and, by its nature, with more context as SMANLIs rationale generation was solely based on a single part of the data point (i.e., the premise). This semantic depth aims to enforce the SLM we fine-tune to reason about its label choice. A prompt was constructed as follows:
\begin{lstlisting}[language=Python]
"Explain why this text 
{premise}
entails the following hypothesis:
{hypothesis}
in 100 words or less."
\end{lstlisting}
Line 3 of the prompt was replaced by either "is neutral to the following hypothesis: " for the neutral label or "contradicts the following hypothesis" for the contradiction label.

\subsection{Model Training and Evaluation}
\begin{figure}
    \centering
    \includegraphics[width=\textwidth]{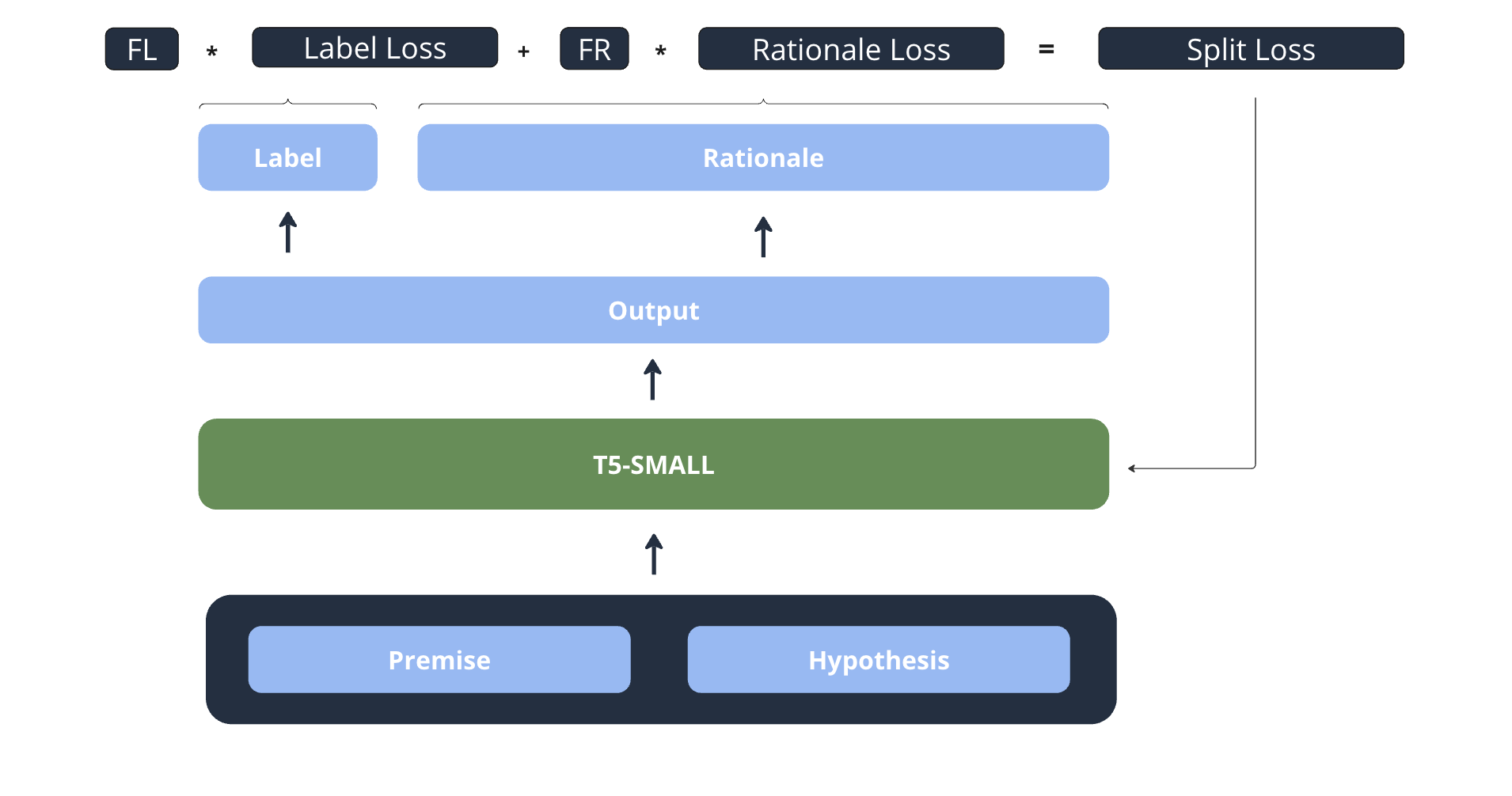}
    \caption{Fine-Tuning process of T5-Small.}
    \label{fig:fine-tuning}
\end{figure}
To assess the impact of these artificially augmented NLI datasets on a student model's performance on label classification, multiple versions of T5-Small were fine-tuned. The training process was implemented using the Hugginface API and distributed on two NVIDIA RTX A6000 GPUs for five epochs, conducting two evaluation- and saving-steps per epoch.

We monitored the overall loss, label classification accuracy, and rationale score using the simple Bilingual Evaluation Understudy (BLEU) \cite{papineni2002bleu} measure for textual similarity. This was measured mainly to gain insights into the training progress and split ratio behavior.

A T5-Small without rationales was fine-tuned to serve as a baseline. For the I$\to$OR models, we applied a custom split-loss defined in equation \ref{equ:1} to make use of the text-to-text architecture while maintaining adequate focus on the label.

Our training pipeline can be found in Figure~\ref{fig:fine-tuning}.
The custom loss is computed by adding factorized Cross-Entropy-Losses for the label and the remaining output of the model (i.e. the rationale) respectively, where the different split ratio in Table \ref{table:categorical_structure} define the corresponding loss fractions as 

\begin{equation}
    \textit{SplitRatio} = (F_{\textit{Label}}, F_{\textit{Rationale}}), \text{  where  } F_{\textit{Label}} + F_{\textit{Rationale}} = 1
\end{equation}

Equation \ref{equ:1} defines the split loss as the sum of two cross-entropy losses, each calculated from distinct data —label-related loss (\(y_{l,c}\) and \(p_{l,c}\)) and rationale-related loss (\(y_{r,c}\) and \(p_{r,c}\)), respectively. They are weighted by their corresponding factors (\(F_{\textit{Label}}\) and \(F_{\textit{Rationale}}\)).
\begin{equation}
    \textit{Loss}_{\textit{split}}  = F_{\textit{Label}} * -\sum_{c=1}^M y_{l,c}\log(p_{l,c}) + F_{\textit{Rationale}} * -\sum_{c=1}^M y_{r,c}\log(p_{r,c}) 
    \label{equ:1}
\end{equation}

Due to the increased uncertainty of hyperparameters induced by the custom loss and the specific goal of label accuracy, a variety of model configurations were fine-tuned on each dataset (see Table \ref{table:categorical_structure}). A combination of learning rate adjustment and split ratio for the loss was explored to extract the optimal configuration for optimal classification accuracy during testing. 

For selecting the final model, we evaluated two separate checkpoints for each model. One based on smallest evaluation loss and one based highest label accuracy to counter performance differences on the test dataset due to the differently implemented loss functions.

\begin{table}[h]
\centering
\caption{Tested T5-Small configurations of split ratios and learning rates.}
\label{table:categorical_structure}
\begin{tabular}{|l|l|l|l|l|}
\hline
 & \textbf{Split Ratio $\to$ Learning Rate} & \textbf{6e-4} & \textbf{1.2e-3} & \textbf{2.4e-3} \\ \hline
Baseline & None & B1 & B2 & B3 \\ \hline
\multirow{3}{*}{T5 SMANLI} & (0.25, 0.75) & S1 & S2 & S3 \\ \cline{2-5} 
 & (0.5, 0.5) & S4 & S5 & S6 \\ \cline{2-5} 
 & (0.75, 0.25) & S7 & S8 & S9 \\ \hline
\multirow{3}{*}{T5 EMANLI} & (0.25, 0.75) & E1 & E2 & E3 \\ \cline{2-5} 
 & (0.5, 0.5) & E4 & E5 & E6 \\ \cline{2-5} 
 & (0.75, 0.25) & E7 & E8 & E9 \\ \hline
\end{tabular}
\end{table}

\section{Results}

\subsection{Enhanced Performance with Rationale Integration}
\begin{figure}
\includegraphics[width=\textwidth]{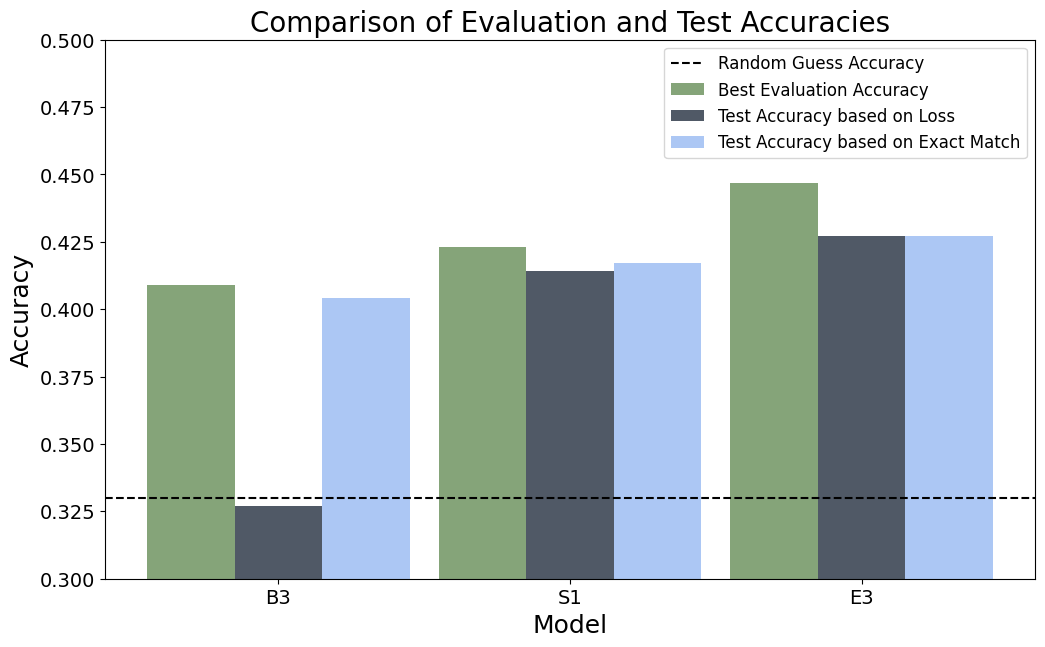}
\caption{Results of the overall best models on tests data for each category of NLI dataset.} 
\label{fig:bestModels}
\end{figure}
\begin{figure}[t!]
\includegraphics[width=\textwidth]{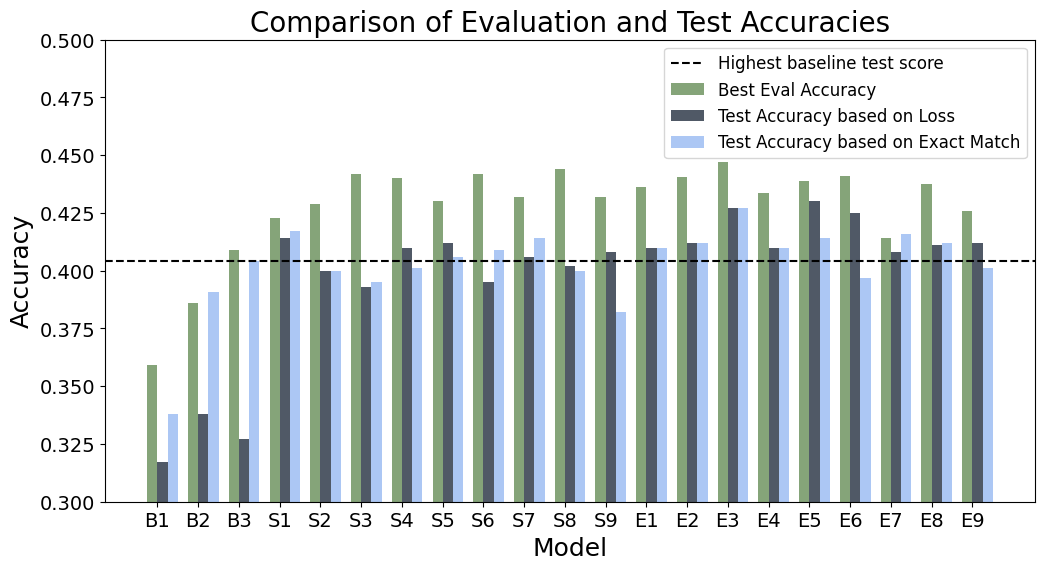}
\caption{Overview of all results of T5-Small model configurations.} 
\label{fig:overviewResults}
\end{figure}

Our results, as shown in Figure \ref{fig:bestModels}, provide evidence for the fact that synthetically augmenting the ANLI dataset using ChatGPT-3.5-Turbo significantly enhances T5-Smalls performance while keeping the augmentation cost minimal. T5-Small trained on SMANLI achieved a test score of 41.7\% (1.3\% over baseline). The T5-Small fine-tuned on EMANLI showed an even stronger performance of 42.7\% on both tests, exceeding the baseline by 2.3\%. These results show that without human intervention or generation, LLMs can act as a teacher model for a very small model through the model-independent dataset augmentation procedure.
An overview of the results of all model configurations is provided in Table \ref{table:categorical_structure} and figure~\ref{fig:overviewResults}.

\subsection{Consistency and Training Efficiency}
\begin{figure}[htbp]
     \centering
     \begin{subfigure}[b]{0.49\textwidth}
         \centering
         \includegraphics[width=\textwidth]{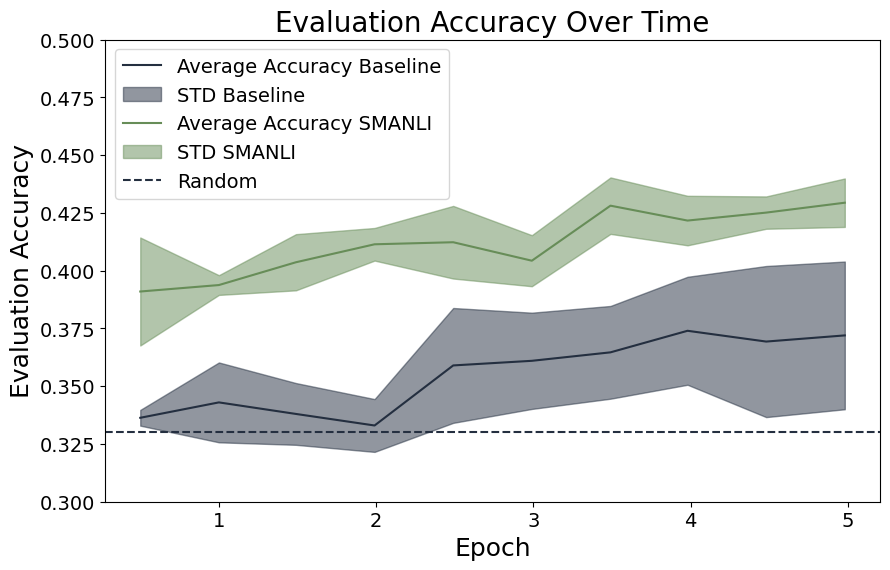}
         \caption{T5-Small trained on SMANLI.}
         \label{fig:T5-Small trained on SMANLI}
     \end{subfigure}
     \hfill
     \begin{subfigure}[b]{0.49\textwidth}
         \centering
         \includegraphics[width=\textwidth]{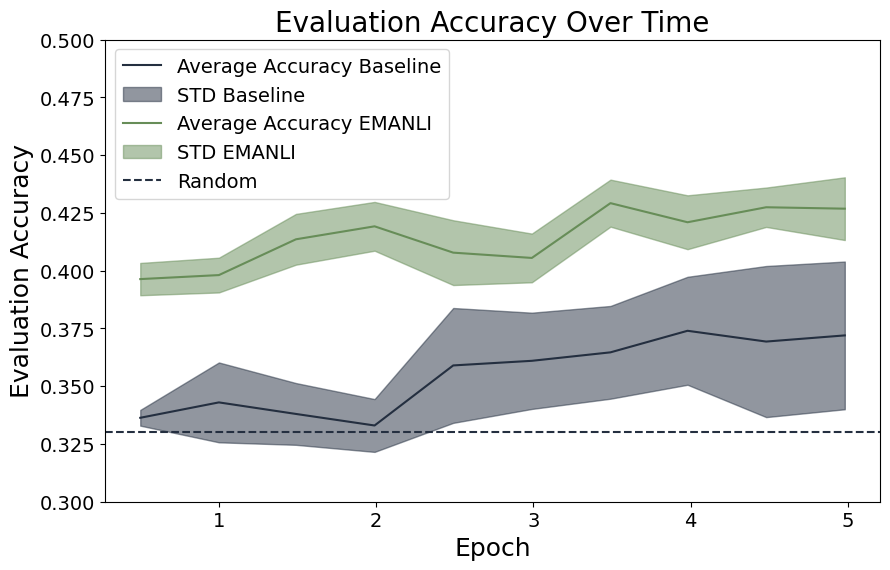}
         \caption{T5-Small trained on EMANLI.}
         \label{fig:T5-Small trained on EMANLI}
     \end{subfigure}

        \caption{Evaluation Accuracy during training process.}
        \label{fig:Evaluation Accuracy during training process}
\end{figure}

During the training process, models generating rationales in addition to their output (I$\to$OR) not only achieved higher test accuracies on most configurations but also exhibited magnitudes higher starting accuracies at the beginning of training. This suggests that rationale generation could indeed enhance the understanding of SLMs like T5-Small, leading to improved training efficiency and, hence, possibly shorter, more efficient fine-tuning times. The accuracies on the evaluation dataset can be found in Figure \ref{fig:Evaluation Accuracy during training process}.

\subsection{Rationale Quality and Model Performance}
\begin{figure}[t]
     \centering
     \begin{subfigure}[b]{0.49\textwidth}
         \centering
         \includegraphics[width=\textwidth]{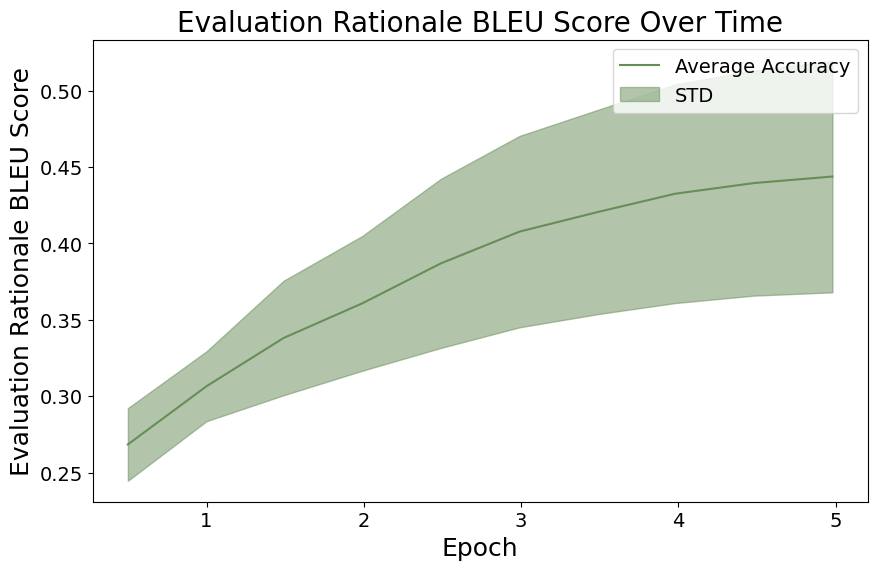}
         \caption{T5-Small trained on SMANLI.}
         \label{fig:RationaleQualitySMANLI}
     \end{subfigure}
     \hfill
     \begin{subfigure}[b]{0.49\textwidth}
         \centering
         \includegraphics[width=\textwidth]{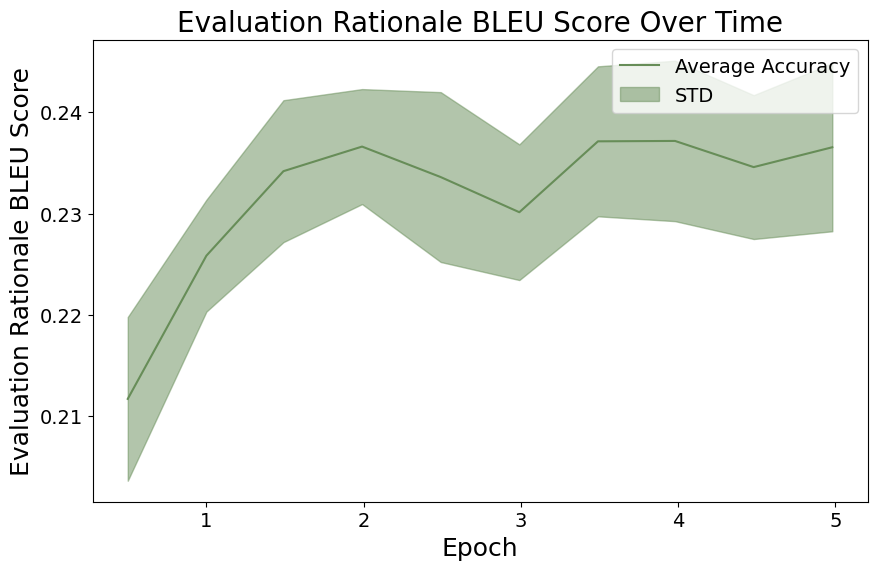}
         \caption{T5-Small trained on EMANLI.}
         \label{fig:RationaleQualityEMANLI}
     \end{subfigure}

        \caption{Evaluation Accuracy during training process.}
        \label{fig:rationaleScore}
\end{figure}
The variations in BLEU score between the two versions of the dataset, as well as the progression of the score throughout the training phase (Figure \ref{fig:rationaleScore}), reveal the core differences in the impact of augmenting ANLI with different types of rationales on the training process. These variations, specifically, highlight how such augmentations influence the label accuracy of the T5-Small model.

While SMANLI achieves an overall much higher rationale score -- originating from the smaller possible variance in information extraction in comparison to free-text explanations -- the development of the scores differs during training.
The rationale score for EMANLI shows a direct correlation with the development of label accuracy during training, while the SMANLI rationale score steadily increases.

We, therefore, conclude that for information extraction tasks, although there is an improvement present, the models simply learn additional output, whereas, for free-text explanations, the quality (in terms of similarity to the teacher's explanation) directly influences the model's capability and score of label classification.

\section{Conclusion and Future Work}
Our results affirm the increase of potential in small language models, given that artificially generated rationales by a teacher model are used in the fashion of knowledge distillation through dataset augmentation. By leveraging the teacher model's capabilities and comprehensive skills, we enhanced the performance of a student model and its reliability and consistency under various factors in an extremely cost-efficient and fast manner.

Additionally, we gave insight into the behavior and effect of two different rationale concepts and showed how information extraction and informed reasoning influenced rationale generation and label accuracy.
These findings expand on the field of dataset synthesis and open new avenues in different sub-fields. 
We, therefore, suggest further deep dives, optimization, and expansion of the approach.
These include testing and comparing more advanced models like FlanT5 \cite{wei2022flan2}, differently sized models (e.g. T5-Base or other counterparts), as well as different teacher models or framing of the prompts.
Additionally, a deeper dive into the rationales output by the T5 should be done to obtain insights into the model's explainability as well as correlation towards the label accuracy and overall performance.
Finally, the quality of the augmented dataset was purely evaluated by the analysis of its effects on the student model's performance. We therefore suggest a deeper analysis of the generated data itself to exclude potential biases and limitations for future research.

\newpage
\bibliographystyle{splncs04}
\bibliography{ref.bib}
\end{document}